\def\year{2018}\relax
\documentclass[letterpaper]{article} 
\usepackage{aaai18}  
\usepackage{times}  
\usepackage{helvet}  
\usepackage{courier}  
\usepackage{url}  
\usepackage{graphicx}  

\usepackage{subfigure}
\usepackage{amsmath}
\usepackage{multirow}
\usepackage{arydshln}
\usepackage{color}
\usepackage{amsmath,amssymb,bm}
\usepackage{algorithm}
\usepackage{algorithmic}
\usepackage[american]{babel}
\usepackage[c]{esvect}
\usepackage{footnote}

\def\newcite#1{\citeauthor{#1}~\shortcite{#1}}

\def\newcite#1{\citeauthor{#1}~\shortcite{#1}}

\makeatletter
\newcommand\argmax{\mathop{\operator@font arg~max}}
\newcommand\argmin{\mathop{\operator@font arg~min}}
\makeatother

\begin{document}
\title{Adversarial Learning for Chinese NER from Crowd Annotations\thanks{The corresponding author is Wenliang Chen.} }
\author{Yaosheng  Yang\textsuperscript{1}, Meishan  Zhang\textsuperscript{4}, Wenliang Chen\textsuperscript{1} \\
      {\bf \Large Wei Zhang\textsuperscript{2}, Haofen Wang\textsuperscript{3}, Min Zhang\textsuperscript{1}} \\
\textsuperscript{1}School of Computer Science and Technology, Soochow University, China\\
\textsuperscript{2}Alibaba Group \and \textsuperscript{3}Shenzhen Gowild Robotics Co. Ltd \\
\textsuperscript{4}School of Computer Science and Technology, Heilongjiang University, China\\
\textsuperscript{1}ysyang@stu.suda.edu.cn, \{wlchen, minzhang\}@suda.edu.cn\\
\textsuperscript{4}mason.zms@gmail.com, \textsuperscript{2}lantu.zw@alibaba-inc.com, \textsuperscript{3}wang\_haofen@gowild.cn\\
}
\maketitle
\begin{abstract}
To quickly obtain new labeled data, we can choose crowdsourcing as an alternative way at lower cost in a short time. But as an exchange, crowd annotations from non-experts may be of lower quality than those from experts. In this paper, we propose an approach to performing crowd annotation learning for Chinese Named Entity Recognition (NER) to make full use of the noisy sequence labels from multiple annotators. Inspired by adversarial learning, our approach uses a common Bi-LSTM and a private Bi-LSTM for representing annotator-generic and -specific information. The annotator-generic information is the common knowledge for entities easily mastered by the crowd. Finally, we build our Chinese NE tagger based on the LSTM-CRF model. In our experiments, we create two data sets for Chinese NER tasks from two domains. The experimental results show that our system achieves better scores than strong baseline systems.

\end{abstract}

\section{Introduction}

There has been significant progress on Named Entity Recognition (NER) in recent years using models based on machine learning algorithms \cite{zhao2008unsupervised,collobert2011natural,lample-EtAl:2016:N16-1}. As with other Natural Language Processing (NLP) tasks, building NER systems typically requires a massive amount of labeled training data which are annotated by experts. In real applications, we often need to consider new types of entities in new domains where we do not have existing annotated data. For such new types of entities, however, it is very hard to find experts to annotate the data within short time limits and hiring experts is costly and non-scalable, both in terms of time and money.

In order to quickly obtain new training data, we can use crowdsourcing as one alternative way at lower cost in a short time. But as an exchange, crowd annotations from non-experts may be of lower quality than those from experts.
It is one biggest challenge to build a powerful NER system on such a low quality annotated data.
Although we can obtain high quality annotations for each input sentence by majority voting,
it can be a waste of human labors to achieve such a goal,
especially for some ambiguous sentences which may require a number of annotations to reach an agreement.
Thus majority work directly build models on crowd annotations,
trying to model the differences among annotators,
for example, some of the annotators may be more trustful \cite{rodrigues2014sequence,nguyen2017aggregating}.

Here we focus mainly on the Chinese NER,
which is more difficult than NER for other languages such as English for the lack of morphological variations such as capitalization and in particular the uncertainty in word segmentation.
The Chinese NE taggers trained on news domain often perform poor in other domains.
Although we can alleviate the problem by using character-level tagging to resolve the problem of poor word segmentation  performances \cite{peng-dredze:2015:EMNLP},
still there exists a large gap when the target domain changes, especially for the texts of social media.
Thus, in order to get a good tagger for new domains and also for the conditions of new entity types,
we require large amounts of labeled data.
Therefore, crowdsourcing is a reasonable solution for these situations.

In this paper, we propose an approach to training a Chinese NER system on the crowd-annotated data.
Our goal is to extract additional annotator independent features by adversarial training,
alleviating the annotation noises of non-experts.
The idea of adversarial training in neural networks has been used successfully in several NLP tasks,
such as cross-lingual POS tagging \cite{kim-EtAl:2017:EMNLP2017} and cross-domain POS tagging \cite{gui-EtAl:2017:EMNLP20172}.
They use it to reduce the negative influences of the input divergences among different domains or languages, while we use adversarial training to reduce the negative influences brought by different crowd annotators.
To our best knowledge, we are the first to apply adversarial training for crowd annotation learning.

In the learning framework, we perform adversarial training between the basic NER and an additional worker discriminator.
We have a common Bi-LSTM for representing annotator-generic information and a private Bi-LSTM for representing annotator-specific information.
We build another label Bi-LSTM by the crowd-annotated NE label sequence which reflects the mind of the crowd annotators who learn entity definitions by reading the annotation guidebook.
The common and private Bi-LSTMs are used for NER,
while the common and label Bi-LSTMs are used as  inputs for the worker discriminator.
The parameters of the common Bi-LSTM are learned by adversarial training,
maximizing the worker discriminator loss and meanwhile minimizing the NER loss.
Thus the resulting features of the common Bi-LSTM are worker invariant and NER sensitive.

For evaluation, we create two Chinese NER datasets in two domains: dialog and e-commerce.
We require the crowd annotators to label the types of entities, including person, song, brand, product, and so on.
Identifying these entities is useful for chatbot and e-commerce platforms \cite{kluwer2011chatbots}.
Then we conduct experiments on the newly created datasets to verify the effectiveness of the proposed adversarial neural network model.
The results show that our system outperforms very strong baseline systems.
In summary, we make the following contributions:
\begin{itemize}
  \item We propose a crowd-annotation learning model based on adversarial neural networks. The model uses labeled data created by non-experts to train a NER classifier and simultaneously learns the common and private features among the non-expert annotators.
  \item We create two data sets in dialog and e-commerce domains by crowd annotations. The experimental results show that the proposed approach performs the best among all the comparison systems.

\end{itemize}

\section{Related Work}
Our work is related to three lines of research: Sequence labeling, Adversarial training, and Crowdsourcing.

\noindent\bfseries Sequence labeling.
\mdseries    NER is widely treated as a sequence labeling problem,
by assigning a unique label over each sentential word \cite{ratinov-roth:2009:CoNLL}.
Early studies on sequence labeling often use the models of HMM, MEMM, and CRF \cite{lafferty2001conditional} based on manually-crafted discrete features,
which can suffer the feature sparsity problem and require heavy feature engineering.
Recently, neural network models have been successfully applied to sequence labeling \cite{collobert2011natural,huang2015bidirectional,lample-EtAl:2016:N16-1}.
Among these work,  the model which uses Bi-LSTM for feature extraction and CRF for decoding has achieved state-of-the-art performances \cite{huang2015bidirectional,lample-EtAl:2016:N16-1},
which is exploited as the baseline model in our work.

\noindent\bfseries Adversarial Training.  \mdseries   Adversarial Networks have achieved great success in computer vision such as image generation \cite{denton2015deep,ganin2016domain}.
In the NLP community, the method is mainly exploited under the settings of domain adaption \cite{zhang2017aspect,gui-EtAl:2017:EMNLP20172}, cross-lingual \cite{chen2016adversarial,kim-EtAl:2017:EMNLP2017}
and multi-task learning \cite{chen2017adversarial,liu-qiu-huang:2017:Long}.
All these settings involve the feature divergences between the training and test examples,
and aim to learn invariant features across the divergences
by an additional adversarial discriminator, such as domain discriminator.
Our work is similar to these work but is applies on crowdsourcing learning,
aiming to find invariant features among different crowdsourcing workers.

\noindent\bfseries Crowdsourcing.  \mdseries Most NLP tasks require a massive amount of labeled training data which are annotated by experts. However, hiring experts is costly and non-scalable, both in terms of time and money. Instead, crowdsourcing is another solution to obtain labeled data at a lower cost but with relative lower quality than those from experts. \newcite{snow2008cheap} collected labeled results for several NLP tasks from Amazon Mechanical Turk and demonstrated that non-experts annotations were quite useful for training new systems. In recent years, a series of work have focused on how to use crowdsourcing data efficiently in tasks such as classification \cite{felt2015early,bi2014learning}, and compare quality of crowd and expert labels \cite{dumitrache2017crowdsourcing}.

In sequence labeling tasks, \newcite{dredze2009sequence} viewed this task as a multi-label problem while \newcite{rodrigues2014sequence} took workers identities into account by assuming that each sentential word was tagged correctly by one of the crowdsourcing workers and proposed a CRF-based model with multiple annotators. \newcite{nguyen2017aggregating} introduced a crowd representation in which the
crowd vectors were added into the LSTM-CRF model at train time, but ignored them at test time.
In this paper, we apply adversarial training on crowd annotations on Chinese NER in new domains,
and achieve better performances than previous studies on crowdsourcing learning.

\begin{figure*}[tb]
\centerline{\includegraphics[scale=0.7]{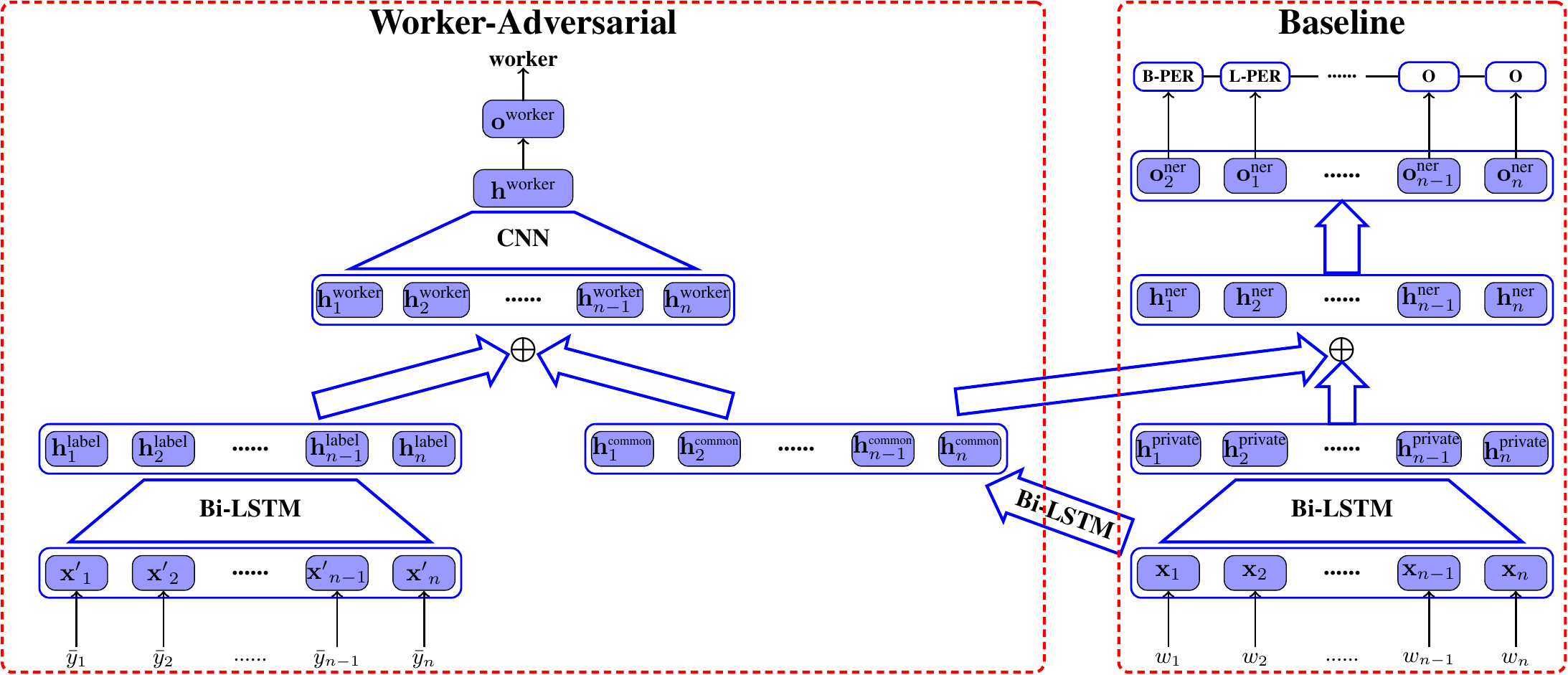}}
\caption{The framework of the proposed model, which consists of two parts.}
\label{fig:model:framework}
\end{figure*}

\section{Baseline: LSTM-CRF}
We use a neural CRF model as the baseline system \cite{ratinov-roth:2009:CoNLL},
treating NER as a sequence labeling problem over Chinese characters,
which has achieved state-of-the-art performances \cite{peng-dredze:2015:EMNLP}.
To this end, we explore the BIEO schema to convert NER into sequence labeling,
following \newcite{lample-EtAl:2016:N16-1},
where sentential character is assigned with one unique tag.
Concretely, we tag the non-entity character by label ``O'',
the beginning character of an entity by ``B-XX'',
the ending character of an entity by ``E-XX'' and the other character of an entity by ``I-XX'',
where ``XX'' denotes the entity type.

We build high-level neural features from the input character sequence by a bi-directional LSTM \cite{lample-EtAl:2016:N16-1}.
The resulting features are combined and then are fed into an output CRF layer for decoding.
In summary, the baseline model has three main components.
First, we make vector representations for sentential characters $\mathbf{x}_1\mathbf{x}_2\cdots \mathbf{x}_n$,
transforming the discrete inputs into low-dimensional neural inputs.
Second, feature extraction is performed to obtain high-level features $\mathbf{h}_1^{\text{ner}}\mathbf{h}_2^{\text{ner}}\cdots \mathbf{h}_n^{\text{ner}}$, by using a bi-directional LSTM (Bi-LSTM)  structure together with a linear transformation over $\mathbf{x}_1\mathbf{x}_2\cdots \mathbf{x}_n$.
Third, we apply a CRF tagging module over $\mathbf{h}_1^{\text{ner}}\mathbf{h}_2^{\text{ner}}\cdots \mathbf{h}_n^{\text{ner}}$, obtaining the final output NE labels.
The overall framework of the baseline model is shown by the right part of Figure \ref{fig:model:framework}.

\subsection{Vector Representation of Characters}
To represent Chinese characters,
we simply exploit a neural embedding layer to map discrete characters into the low-dimensional vector representations.
The goal is achieved by a looking-up table $\mathbf{E}^W$,
which is a model parameter and will be fine-tuned during training.
The looking-up table can be initialized either by random or by using a pretrained embeddings from large scale raw corpus.
For a given Chinese character sequence $c_1c_2\cdots c_n$,
we obtain the vector representation of each sentential character by:
 $ \mathbf{x}_t = \text{look-up}(c_t, \mathbf{E}^W), \text{~~~} t \in [1, n]$.

\subsection{Feature Extraction}
Based on the vector sequence $\mathbf{x}_1\mathbf{x}_2\cdots \mathbf{x}_n$,
we extract higher-level features $\mathbf{h}_1^{\text{ner}}\mathbf{h}_2^{\text{ner}}\cdots \mathbf{h}_n^{\text{ner}}$ by using a bidirectional LSTM module and a simple feed-forward neural layer,
which are then used for CRF tagging at the next step.

LSTM is a type of recurrent neural network (RNN),
which is designed for solving the exploding and diminishing gradients of basic RNNs \cite{graves2005framewise}.
It has been widely used in a number of NLP tasks,
including POS-tagging \cite{huang2015bidirectional,ma-hovy:2016:P16-1}, parsing \cite{dyer-EtAl:2015:ACL-IJCNLP} and machine translation \cite{wu2016google},
because of its strong capabilities of modeling natural language sentences.

By traversing $\mathbf{x}_1\mathbf{x}_2\cdots \mathbf{x}_n$  by order and reversely,
we obtain the output features $\mathbf{h}_1^{\text{private}}\mathbf{h}_2^{\text{private}}\cdots \mathbf{h}_n^{\text{private}}$ of the bi-LSTM,
where $\mathbf{h}_t^{\text{private}} = \overrightarrow{\mathbf{h}}_t \oplus  \overleftarrow{\mathbf{h}}_t $.
Here we refer this Bi-LSTM as private in order to differentiate it with the common Bi-LSTM over the same character inputs which will be introduced in the next section.

Further we make an integration of the output vectors of bi-directional LSTM by a linear feed-forward neural layer,
resulting in the features  $\mathbf{h}_1^{\text{ner}}\mathbf{h}_2^{\text{ner}}\cdots \mathbf{h}_n^{\text{ner}}$ by equation:
\begin{equation}\label{base-combine}
 \mathbf{h}_t^{\text{ner}} = \mathbf{W} \mathbf{h}_t^{\text{private}} + \mathbf{b},
\end{equation}
where $\mathbf{W}$ and $\mathbf{b}$ are both model parameters.

\subsection{CRF Tagging}
Finally we feed the resulting features $\mathbf{h}_t^{\text{ner}}, t\in [1, n]$ into a CRF layer directly for NER decoding.
CRF tagging is one globally normalized model,
aiming to find the best output sequence considering the dependencies between successive labels.
In the sequence labeling setting for NER,
the output label of one position has a strong dependency on the label of the previous position.
For example, the label before ``I-XX'' must be either ``B-XX'' or ``I-XX'',
where ``XX'' should be exactly the same.

CRF involves two parts for prediction.
First we should compute the scores for each label based $\mathbf{h}_t^{\text{ner}}$,
resulting in $\mathbf{o}_t^{\text{ner}}$,
whose dimension is the number of output labels.
The other part is a transition matrix $\mathbf{T}$
which defines the scores of two successive labels.
$\mathbf{T}$ is also a model parameter.
Based on $\mathbf{o}_t^{\text{ner}}$ and $\mathbf{T}$,
we use the Viterbi algorithm to find the best-scoring label sequence.

We can formalize the CRF tagging process as follows:
\begin{equation}\label{output-ner}
\begin{split}
& \mathbf{o}_t^{\text{ner}}  = \mathbf{W}^{\text{ner}} \mathbf{h}_t^{\text{ner}}, \text{~~~~} t \in [1,n] \\
& \text{score}(\mathbf{X}, \mathbf{y}) = \sum_{t = 1}^{n}(\mathbf{o}_{t,y_t} + T_{y_{t-1},y_t}) \\
& \mathbf{y}^{\text{ner}} = \argmax_{\mathbf{y}}\big(\text{score}(\mathbf{X}, \mathbf{y}))\big), \\
\end{split}
\end{equation}
where $\text{score}(\cdot)$ is the scoring function
for a given output label sequence $\mathbf{y} = y_1y_2 \cdots y_n$ based on input $\mathbf{X}$,
$\mathbf{y}^{\text{ner}}$ is the resulting label sequence, $\mathbf{W}^{\text{ner}}$ is a model parameter.

\subsection{Training}
To train model parameters,
we exploit a negative log-likelihood objective as the loss function.
We apply softmax over all candidate output label sequences,
thus the probability of the crowd-annotated label sequence is computed by:
\begin{equation}\label{crf-prob}
p(\mathbf{\bar{y}}|\mathbf{X}) = \frac{\exp\big(\text{score}(\mathbf{X}, \mathbf{\bar{y}})\big)} {\sum_{\mathbf{y} \in \mathbf{Y}_{\mathbf{X}}} \exp\big(\text{score}(\mathbf{X}, \mathbf{y})\big)},
\end{equation}
where $\mathbf{\bar{y}}$ is the crowd-annotated label sequences and $\mathbf{Y}_{\mathbf{X}}$ is all candidate label sequence of input $\mathbf{X}$.

Based on the above formula, the loss function of our baseline model is:
\begin{equation}\label{crf-obj}
\text{loss}(\Theta, \mathbf{X}, \mathbf{\bar{y}}) = -\log p(\mathbf{\bar{y}}|\mathbf{X}),
\end{equation}
where $\Theta$ is the set of all model parameters.
We use standard back-propagation method to minimize the loss function of the baseline CRF model.

\section{Worker Adversarial}
Adversarial learning has been an effective mechanism to resolve the problem of
the input features between the training and test examples having large divergences  \cite{goodfellow2014generative,ganin2016domain}.
It has been successfully applied on domain adaption \cite{gui-EtAl:2017:EMNLP20172}, cross-lingual learning \cite{chen2016adversarial} and  multi-task learning \cite{liu-qiu-huang:2017:Long}.
All settings involve feature shifting between the training and testing.

In this paper, our setting is different.
We are using the annotations from non-experts,
which are noise and can influence the final performances if they are not properly processed.
Directly learning based on the resulting corpus may adapt the neural feature extraction into the biased annotations.
In this work,
we assume that individual workers have their own guidelines in mind after short training.
For example, a perfect worker can annotate highly consistently with an expert,
while common crowdsourcing workers may be confused and have different understandings on certain contexts.
Based on the assumption,
we make an adaption for the original adversarial neural network to our setting.

Our adaption is very simple.
Briefly speaking, the original adversarial learning adds an additional discriminator to classify
the type of source inputs, for example, the domain category in the domain adaption setting,
while we add a discriminator to classify the annotation workers.
Solely the features from the input sentence is not enough for worker classification.
The annotation result of the worker is also required.
Thus the inputs of our discriminator are different.
Here we exploit both the source sentences and the crowd-annotated NE labels as basic inputs for the worker discrimination.

In the following, we describe the proposed adversarial learning module,
including both the submodels and the training method.
As shown by the left part of Figure \ref{fig:model:framework},
the submodel consists of four parts: (1) a common Bi-LSTM over input characters;
(2) an additional Bi-LSTM to encode crowd-annotated NE label sequence;
(3) a convolutional neural network (CNN) to extract features for worker discriminator;
(4) output and prediction.

\subsection{Common Bi-LSTM over Characters}
To build the adversarial part,
first we create a new bi-directional LSTM, named by the common Bi-LSTM:
\begin{equation}\label{LSTM-common}
\mathbf{h}_1^{\text{\tiny common}} \mathbf{h}_2^{\text{\tiny common}} \cdots \mathbf{h}_n^{\text{\tiny common}} = \text{Bi-LSTM}(\mathbf{x}_1\mathbf{x}_2\cdots \mathbf{x}_n).
\end{equation}
As shown in Figure \ref{fig:model:framework}, this Bi-LSTM is constructed over the same input character representations of the private Bi-LSTM, in order to extract worker independent features.

The resulting features of the common Bi-LSTM are used for both NER and the worker discriminator,
different with the features of private Bi-LSTM which are used for NER only.
As shown in Figure \ref{fig:model:framework},
we concatenate the outputs of the common and private Bi-LSTMs together,
and then feed the results into the feed-forward combination layer of the NER part.
Thus Formula \ref{base-combine} can be rewritten as:
\begin{equation}\label{new-combine}
 \mathbf{h}_t^{\text{ner}} = \mathbf{W} (\mathbf{h}_t^{\text{common}} \oplus \mathbf{h}_t^{\text{private}}) + \mathbf{b},
\end{equation}
where $\mathbf{W}$ is wider than the original combination because the newly-added $\mathbf{h}_t^{\text{common}}$.

Noticeably, although the resulting common features are used for the worker discriminator,
they actually have no capability to distinguish the workers.
Because this part is exploited to maximize the loss of the worker discriminator,
it will be interpreted in the later training subsection.
These features are invariant among different workers,
thus they can have less noises for NER.
This is the goal of adversarial learning,
and we hope the NER being able to find useful features from these worker independent features.

\subsection{Additional Bi-LSTM over Annotated NER Labels}
In order to incorporate the annotated NE labels to predict the exact worker,
we build another bi-directional LSTM (named by label Bi-LSTM) based on the crowd-annotated NE label sequence.
This Bi-LSTM is used for worker discriminator only.
During the decoding of the testing phase,
we will never have this Bi-LSTM, because the worker discriminator is no longer required.

Assuming the crowd-annotated NE label sequence annotated by one worker is $\mathbf{\bar{y}} = \bar{y}_1\bar{y}_2 \cdots \bar{y}_n$,
we exploit a looking-up table $\mathbf{E}^{L}$ to obtain the corresponding sequence of their vector representations $\mathbf{x'}_1\mathbf{x'}_2\cdots \mathbf{x'}_n$,
similar to the method that maps characters into their neural representations.
Concretely, for one NE label $\bar{y}_t$ ($t \in [1, n]$),
we obtain its neural vector by:
$\mathbf{x'}_t = \text{look-up}(\bar{y}_t, \mathbf{E}^L)$.

Next step we apply bi-directional LSTM over the sequence $\mathbf{x'}_1\mathbf{x'}_2\cdots \mathbf{x'}_n$,
which can be formalized as:
\begin{equation}\label{label-lstm}
\mathbf{h}_1^{\text{label}} \mathbf{h}_2^{\text{label}} \cdots \mathbf{h}_n^{\text{label}} = \text{Bi-LSTM}(\mathbf{x'}_1\mathbf{x'}_2\cdots \mathbf{x'}_n).
\end{equation}
The resulting feature sequence is concatenated with the outputs of the common Bi-LSTM,
and further be used for worker classification.

\subsection{CNN}
Following, we add a convolutional neural network (CNN) module based on the concatenated outputs of the common Bi-LSTM and the label Bi-LSTM,
to produce the final features for worker discriminator.
A convolutional operator with window size 5 is used,
and then max pooling strategy is applied over the convolution sequence
to obtain the final fixed-dimensional feature vector.
The whole process can be described by the following equations:
\begin{equation}\label{cnn-worker}
\begin{split}
&\mathbf{h}_t^{\text{worker}}  = \mathbf{h}_t^{\text{common}} \oplus \mathbf{h}_t^{\text{label}}  \\
&\mathbf{\tilde{h}}_t^{\text{worker}}  = \tanh(\mathbf{W}^{\text{cnn}}[\mathbf{h}_{t-2}^{\text{worker}}, \mathbf{h}_{t-1}^{\text{worker}}, \cdots, \mathbf{h}_{t+2}^{\text{worker}}]) \\
&\mathbf{h}^{\text{worker}}  = \text{max-pooling}(\mathbf{\tilde{h}}_1^{\text{worker}}\mathbf{\tilde{h}}_2^{\text{worker}} \cdots \mathbf{\tilde{h}}_n^{\text{worker}}) \\
\end{split}
\end{equation}
where $t \in [1,n]$ and $\mathbf{W}^{\text{cnn}}$ is one model parameter.
We exploit zero vector to paddle the out-of-index vectors.


\subsection{Output and Prediction}
After obtaining the final feature vector for the worker discriminator,
we use it to compute the output vector, which scores all the annotation workers.
The score function is  defined by:
\begin{equation}\label{output-worker}
\mathbf{o}^{\text{worker}}  = \mathbf{W}^{\text{worker}} \mathbf{h}^{\text{worker}},
\end{equation}
where $\mathbf{W}^{\text{worker}}$ is one model parameter
and the output dimension equals the number of total non-expert annotators.
The prediction is to find the worker which is responsible for this annotation.

\subsection{Adversarial Training}
The training objective with adversarial neural network is different from the baseline model,
as it includes the extra worker discriminator.
Thus the new objective includes two parts,
one being the negative log-likelihood from NER
which is the same as the baseline,
and the other being the negative the log-likelihood from the worker discriminator.

In order to obtain the negative log-likelihood of the worker discriminator,
we use softmax to compute the probability of the actual worker $\bar{z}$ as well,
which is defined by:
\begin{equation}\label{output-worker}
p(\bar{z}|\mathbf{X}, \mathbf{\bar{y}})  = \frac{\exp(\mathbf{o}^{\text{worker}}_{\bar{z}})} {\sum_{z} \exp(\mathbf{o}^{\text{worker}}_z)},
\end{equation}
where $z$ should enumerate all workers.

Based on the above definition of probability,
our new objective is defined as follows:
\begin{equation}\label{adver-obj}
\begin{split}
\text{R}(\Theta, \Theta', \mathbf{X}, \mathbf{\bar{y}}, \bar{z}) &= \text{loss}(\Theta, \mathbf{X}, \mathbf{\bar{y}}) -  \text{loss}(\Theta, \Theta', \mathbf{X}) \\
\text{~~~~~~} &= -\log p(\mathbf{\bar{y}}|\mathbf{X}) + \log p(\bar{z}|\mathbf{X}, \mathbf{\bar{y}}),
\end{split}
\end{equation}
where $\Theta$ is the set of all model parameters related to NER,
and $\Theta'$ is the set of the remaining parameters which are only related to the worker discriminator,
$\mathbf{X}$, $\mathbf{\bar{y}}$ and $\bar{z}$ are the input sentence, the crowd-annotated NE labels and the corresponding annotator for this annotation, respectively.
It is worth noting that the parameters of the common Bi-LSTM are included in the set of $\Theta$ by definition.

In particular, our goal is not to simply minimize the new objective.
Actually, we aim for a saddle point,
finding the parameters $\Theta$ and $\Theta'$ satisfying the following conditions:
\begin{equation}\label{adver-cond}
\begin{split}
\hat{\Theta} &= \argmin_{\Theta}\text{R}(\Theta, \Theta', \mathbf{X}, \mathbf{\bar{y}}, \bar{z}) \\
\hat{\Theta}' &= \argmax_{\Theta'}\text{R}(\hat{\Theta}, \Theta', \mathbf{X}, \mathbf{\bar{y}}, \bar{z}) \\
\end{split}
\end{equation}
where the first equation aims to find one $\Theta$ that minimizes our new objective $\text{R}(\cdot)$,
and the second equation aims to find one $\Theta'$ maximizing the same objective.

Intuitively, the first equation of Formula \ref{adver-cond} tries to minimize the NER loss,
but at the same time maximize the worker discriminator loss by the shared parameters of the common Bi-LSTM.
Thus the resulting features of common Bi-LSTM actually attempt to hurt the worker discriminator,
which makes these features worker independent since they are unable to distinguish different workers.
The second equation tries to minimize the worker discriminator loss by its own parameter $\Theta'$.

We use the standard back-propagation method to train the model parameters,
the same as the baseline model.
In order to incorporate the term of the argmax part of Formula \ref{adver-cond} ,
we follow the previous work of adversarial training \cite{ganin2016domain,chen2016adversarial,liu-qiu-huang:2017:Long},
by introducing a gradient reverse layer between the common Bi-LSTM and the CNN module,
whose forward does nothing but the backward simply negates the gradients.

\section{Experiments}
\subsection{Data Sets}
With the purpose of obtaining evaluation datasets from crowd annotators, we collect the sentences from two domains: Dialog and E-commerce domain. We hire undergraduate students to annotate the sentences.
They are required to identify the predefined types of entities in the sentences. Together with the guideline document, the annotators are educated some tips in fifteen minutes and also provided with 20 exemplifying sentences.

\noindent\textbf{Labeled Data: DL-PS.} In Dialog domain (DL), we collect raw sentences from a chatbot application. And then we randomly select 20K sentences as our pool and hire 43 students to annotate the sentences. We ask the annotators to label two types of entities: Person-Name and Song-Name. The annotators label the sentences independently.
In particular, each sentence is assigned to three annotators for this data.
Although the setting can be wasteful of labor,
we can use the resulting dataset to test several well-known baselines such as majority voting.

After annotation, we remove some illegal sentences reported by the annotators. Finally, we have 16,948 sentences annotated by the students. Table \ref{tb:datasets} shows the information of annotated data. The average Kappa value among the annotators is 0.6033, indicating that the crowd annotators have moderate agreement on identifying entities on this data.

In order to evaluate the system performances, we create a set of corpus with gold annotations.
Concretely, we randomly select 1,000 sentences from the final dataset and let two experts generate the gold annotations.
Among them, we use 300 sentences as the development set and the remaining 700 as the test set.
The rest sentences with only student annotations are used as the training set.

\begin{table}[t]
  \centering
  {
  \begin{tabular}{l|c|c|c}
  \hline
   &  \#Sent  & AvgLen  & Kappa\\
  \hline
  DL-PS & 16,948   & 9.21 & 0.6033\\
  UC-MT & 2,337    & 34.97 & 0.7437\\
  UC-UQ & 2,300    & 7.69 & 0.7529\\
  \hline
  \end{tabular}
  \caption{Statistics of labeled datasets.}\label{tb:datasets}
  }
\end{table}

\noindent\textbf{Labeled data: EC-MT and EC-UQ.} In E-commerce domain (EC),  we collect raw sentences from two types of texts: one is titles of merchandise entries (EC-MT) and another is user queries (EC-UQ).
The annotators label five types of entities: Brand, Product, Model, Material, and Specification.
These five types of entities are very important for E-commerce platform, for example building knowledge graph of merchandises.
Five students participate the annotations for this domain since the number of sentences is small.
We use the similar strategy as DL-PS to annotate the sentences,
except that only two annotators are assigned for each sentence,
because we aim to test the system performances under very small duplicated annotations.

Finally, we obtain 2,337 sentences for EC-MT and 2,300 for EC-UQ.  Table \ref{tb:datasets} shows the information of annotated results.
Similarly, we produce the development and test datasets for system evaluation,
by randomly selecting 400 sentences and letting two experts to generate the groundtruth annotations.
Among them, we use 100 sentences as the development set and the remaining 300 as the test set.
The rest sentences with only crowdsourcing annotations are used as the training set.

\noindent\textbf{Unlabeled data.}
The vector representations of characters are basic inputs of our baseline and proposed models,
which are obtained by the looking-up table $\mathbf{E}^W$.
As introduced before, we can use pretrained embeddings from large-scale raw corpus to initialize the table.
In order to pretrain the character embeddings,
we use one large-scale unlabeled data from the user-generated content in Internet.
Totally, we obtain a number of 5M sentences.
Finally, we use the tool \texttt{word2vec}\footnote{https://code.google.com/archive/p/word2vec} to pretrain the character embeddings based on the unlabeled dataset in our experiments.

\subsection{Settings}
For evaluation, we use the entity-level metrics of Precision (P), Recall (R), and their F1 value in our experiments,
treating one tagged entity as correct only when it matches the gold entity exactly.

There are several hyper-parameters in the baseline LSTM-CRF and our final models.
We set them empirically by the development performances.
Concretely, we set the dimension size of the character embeddings by 100,
the dimension size of the NE label embeddings by 50,
and the dimension sizes of all the other hidden features by 200.

We exploit online training with a mini-batch size 128 to learn model parameters.
The max-epoch iteration is set by 200,
and the best-epoch model is chosen according to the development performances.
We use RMSprop \cite{tieleman2012lecture} with a learning rate $10^{-3}$ to update model parameters,
and use $l_2$-regularization by a parameter $10^{-5}$.
We adopt the dropout technique to avoid overfitting by a drop value of $0.2$.

\subsection{Comparison Systems}
The proposed approach (henceforward referred to as “ALCrowd”) is compared with the following systems:
\begin{itemize}
  \item CRF: We use the Crfsuite\footnote{http://www.chokkan.org/software/crfsuite/} tool to train a model on the  crowdsourcing labeled data. As for the feature settings, we use the supervised version of \newcite{zhao2008unsupervised}.
  \item CRF-VT: We use the same settings of the CRF system, except that the training data is the voted version,
  whose groundtruths are produced by majority voting at the character level for each annotated sentence.
  \item CRF-MA: The CRF model proposed by \newcite{rodrigues2014sequence}, which uses a prior distributation to model multiple crowdsourcing annotators. We use the source code provided by the authors.
  \item LSTM-CRF: Our baseline system trained on the crowdsourcing labeled data.
  \item LSTM-CRF-VT: Our baseline system trained on the voted corpus, which is the same as CRF-VT.
  \item LSTM-Crowd: The LSTM-CRF model with crowd annotation learning proposed by \newcite{nguyen2017aggregating}. We use the source code provided by the authors.
\end{itemize}
The first three systems are based on the CRF model using traditional handcrafted features, and the last three systems are based on the neural LSTM-CRF model.
Among them, CRF-MA, LSTM-Crowd and our system with adversarial learning (ALCrowd) are based on crowd annotation learning that directly trains the model on the crowd-annotations.
Five systems, including CRF, CRF-MA, LSTM-CRF, LSTM-Crowd, and ALCrowd, are trained on the original version of labeled data, while CRF-VT and LSTM-CRF-VT are trained on the voted version.
Since CRF-VT, CRF-MA and LSTM-CRF-VT all require ground-truth answers for each training sentence, which are difficult to be produced with only two annotations,
we do not apply the three models on the two EC datasets.

\begin{table}[t]
  \centering
  {
  \begin{tabular}{l|c|c|c}
  \hline
Model	& P	& R	& F1	\\
\hline
CRF	&89.48		&70.38		&78.79 \\
CRF-VT	&85.16		&65.07		&73.77 \\
CRF-MA	&72.83		&\bf 90.79		&80.82 \\ \hline
LSTM-CRF & \bf 90.50	&79.97		&84.91 \\
LSTM-CRF-VT & 88.68 & 75.51 & 81.57 \\
LSTM-Crowd &86.40	&83.43		&84.89 \\
ALCrowd &89.56	&82.70		&\bf 85.99 \\
  \hline
  \end{tabular}
  \caption{Main results on the DL-PS data.}\label{tb:resultsDL}
  }
\end{table}

\begin{table}[t]
\begin{center}
\begin{tabular}{c|ccc}
\hline
\multirow{2}{*}{Model} &   \multicolumn{3}{|c}{ Data: EC-MT}   \\ \cline{2-4}
 &  P & R  &  F1   \\ \hline
CRF        & 75.12	& 66.67	& 70.64  \\ \hline
LSTM-CRF   & 75.02	& 72.84	& 73.91   \\
LSTM-Crowd & 73.81	& \bf 75.18	& 74.49    \\
ALCrowd  & \bf 76.33	& 74.00	& \bf 75.15    \\ \hline
\hline
 &     \multicolumn{3}{|c}{ Data: EC-UQ}  \\  \hline
CRF          & 65.45  & 55.33	& 59.96 \\ \hline
LSTM-CRF     & 71.96  & 66.55	& 69.15 \\
LSTM-Crowd   & 67.51  & \bf 71.10	& 69.26  \\
ALCrowd    & \bf 74.72  & 68.60	& \bf 71.53  \\ \hline

\end{tabular}
\caption{Main results on the EC-MT and EC-UQ datasets. } \label{result:ec}
\end{center}
\end{table}

\subsection{Main Results}

In this section, we show the model performances of our proposed crowdsourcing learning system (ALCrowd),
and meanwhile compare it with the other systems mentioned above. Table \ref{tb:resultsDL} shows the experimental results on the DL-PS datasets
and Table \ref{result:ec} shows the experiment results on the EC-MT and EC-UQ datasets, respectively.

The results of CRF and LSTM-CRF mean that the crowd annotation is an alternative solution with low cost for labeling data that could be used for training a NER system even there are some inconsistencies.
Compared with CRF, LSTM-CRF achieves much better performances on all the three data,
showing +6.12 F1 improvement on DL-PS, +4.51 on EC-MT, and +9.19 on EC-UQ.
This indicates that LSTM-CRF is a very strong baseline system,
demonstrating the effectiveness of neural network.

Interestingly, when compared with CRF and LSTM-CRF, CRF-VT and LSTM-CRF-VT trained on the voted version perform worse in the DL-PS dataset. This trend is also mentioned in \newcite{nguyen2017aggregating}.
This fact shows that the majority voting method might be unsuitable for our task.
There are two possible reasons accounting for the observation.
On the one hand, simple character-level voting based on three annotations for each sentence may be still not enough.
In the DL-PS dataset, even with only two predefined entity types,
one character can have nine NE labels.
Thus the majority-voting may be incapable of handling some cases.
While the cost by adding more annotations for each sentence would be greatly increased.
On the other hand, the lost information produced by majority-voting may be important,
at least the ambiguous annotations denote that the input sentence is difficult for NER.
The normal CRF and LSTM-CRF models without discard any annotations can differentiate these difficult contexts through learning.

Three crowd-annotation learning systems provide better performances than their counterpart systems, (CRF-MA VS CRF) and (LSTM-Crowd/ALCrowd VS LSTM-CRF). Compared with the strong baseline LSTM-CRF, ALCrowd shows its advantage with +1.08 F1 improvements on DL-PS, +1.24 on EC-MT, and +2.38 on EC-UQ, respectively.
This indicates that adding the crowd-annotation learning is quite useful for building NER systems.
In addition, ALCrowd also outperforms LSTM-Crowd on all the datasets consistently,
demonstrating the high effectiveness of ALCrowd in extracting worker independent features.
Among all the systems, ALCrowd performs the best, and significantly better than all the other models (the p-value is below $10^{-5}$ by using t-test).
The results indicate that with the help of adversarial training,
our system can learn a better feature representation from crowd annotation.

\begin{figure}[t]
\centerline{\includegraphics[scale=0.8]{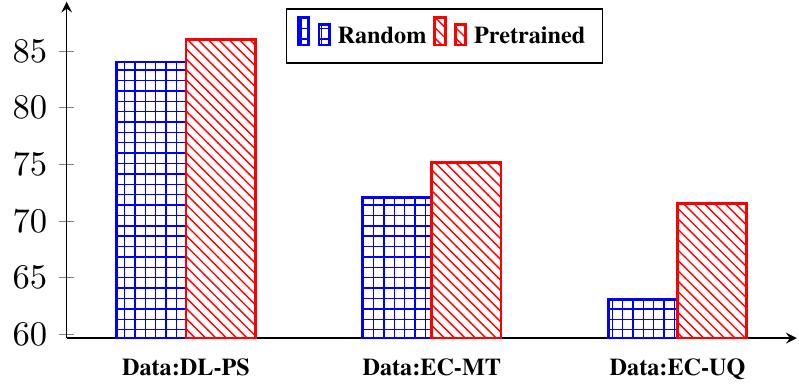}}
\caption{Comparisons by using different character embeddings, where the Y-axis shows the F1 values}
\label{fig:discuss:emb}
\end{figure}

\subsection{Discussion}
\textbf{Impact of Character Embeddings.}
First, we investigate the effect of the pretrained character embeddings in our proposed crowdsourcing learning model.
The comparison results are shown in Figure \ref{fig:discuss:emb},
where \textbf{Random} refers to the random initialized character embeddings,
and \textbf{Pretrained} refers to the embeddings pretrained on the unlabeled data.
According to the results, we find that our model with the pretrained embeddings significantly outperforms that using the random embeddings,
demonstrating that the pretrained embeddings successfully provide useful information.


\begin{figure}[t]
\centerline{\includegraphics[scale=0.3,angle=-90]{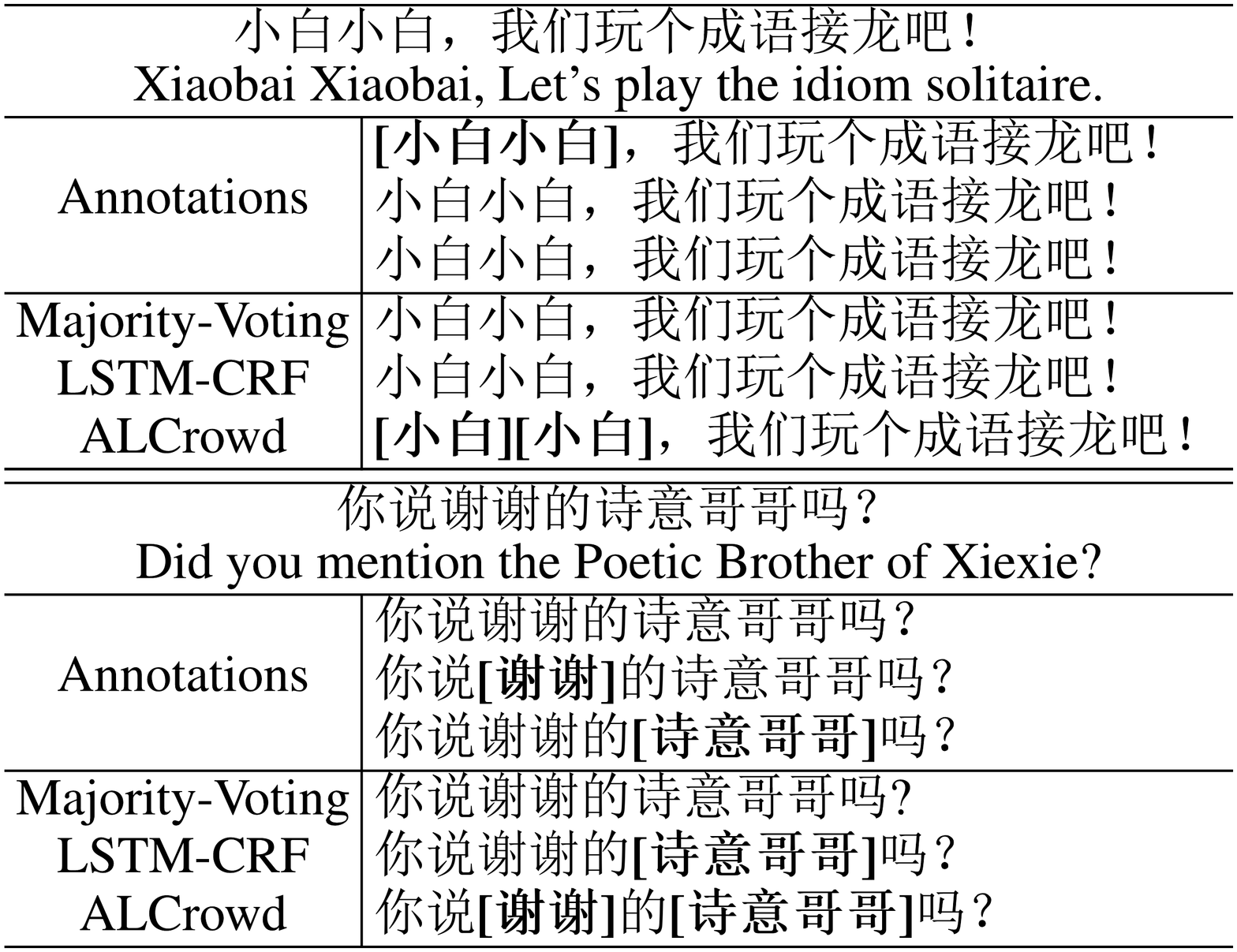}}
\caption{Case studies of different systems, where named entities are illustrated by square brackets.}
\label{tab:case-example}
\end{figure}

\noindent\textbf{Case Studies.}
Second, we present several case studies in order to study the differences between our baseline
and the worker adversarial models.
We conduct a closed test on the training set,
the results of which can be regarded as modifications of the training corpus,
since there exist inconsistent annotations for each training sentence among the different workers.
Figure \ref{tab:case-example} shows the two examples from the DL-PS dataset,
which compares the outputs of the baseline and our final models,
as well as the majority-voting strategy.

In the first case, none of the annotations get the correct NER result,
but our proposed model can capture it.
The result of LSTM-CRF is the same as majority-voting.
In the second example, the output of majority-voting is the worst,
which can account for the reason why the same model trained on the voted corpus performs so badly,
as shown in Table \ref{tb:resultsDL}.
The model of LSTM-CRF fails to recognize the named entity ``Xiexie'' because of not trusting the second annotation, treating it as one noise annotation.
Our proposed model is able to recognize it,
because of its ability of extracting worker independent features.


\section{Conclusions}
In this paper, we presented an approach to performing crowd annotation learning based on the idea of adversarial training for Chinese Named Entity Recognition (NER). In our approach, we use a common and private Bi-LSTMs for representing annotator-generic and -specific information, and learn a label Bi-LSTM from the crowd-annotated NE label sequences. Finally, the proposed approach adopts a LSTM-CRF model to perform tagging. In our experiments, we create two data sets for Chinese NER tasks in the dialog and e-commerce domains. The experimental results show that the proposed approach outperforms strong baseline systems. 

\section{ Acknowledgments}
This work is supported by the National Natural Science Foundation of China (Grant No. 61572338, 61525205, and 61602160). This work is also partially supported by the joint research project of Alibaba and Soochow University. Wenliang is also partially supported by Collaborative Innovation Center of Novel Software Technology and Industrialization.

\bibliographystyle{aaai}
\bibliography{ner}
\end{document}